\documentclass{article}
\usepackage{spconf,amsmath,graphicx}
\usepackage{booktabs}
\usepackage{caption}
\usepackage{subcaption}
\usepackage{graphicx}
\usepackage{comment}
\usepackage{amsmath}
\usepackage{amssymb}
\usepackage{url}

\newcommand{\R}{\mathbb{R}}
\title{Robust Latent Representations via \\Cross-Modal Translation and Alignment}
\name{Vandana Rajan$^{1}$, Alessio Brutti$^{2}$, Andrea Cavallaro$^{1}$}
  \address{$^{1}$Centre for Intelligent Sensing, Queen Mary University of London, UK\\
  $^{2}$Fondazione Bruno Kessler, Trento, Italy }
\begin{document}
\ninept
\maketitle
\begin{abstract}
Multi-modal learning relates information across observation modalities of the same physical phenomenon to leverage complementary information. Most multi-modal machine learning methods require that all the modalities used for training are also available for testing. This is a limitation when signals from some modalities are unavailable or severely degraded. To address this limitation, we aim to improve the testing performance of uni-modal systems using multiple modalities {\em during training only}. The proposed multi-modal training framework uses cross-modal translation and correlation-based latent space alignment to improve the representations of a worse performing (or weaker) modality. The translation from the weaker to the better performing (or stronger) modality generates a multi-modal intermediate encoding that is representative of both modalities. This encoding is then correlated with the stronger modality representation in a shared latent space. We validate the proposed framework on the AVEC 2016 dataset (RECOLA) for continuous emotion recognition and show the effectiveness of the framework that achieves state-of-the-art (uni-modal) performance for weaker modalities. 
\end{abstract}
\section{Introduction}
\label{sec:intro}
The term {\em modality} refers to the particular form in which something exists or is experienced or expressed~\cite{baltruvsaitis2018multimodal}. Most physical phenomena we experience consist of multiple modalities; for example, we can {\em see}, {\em hear} and {\em touch} the rain; objects around us may have their own characteristic shape, sound and smell. The information to explain an event is often unevenly spread across the modalities that capture this event. {\em Multi-modal} machine learning uses multiple modalities to model or explain an event, whereas {\em uni-modal} or {\em mono-modal} machine learning uses only one of these modalities~\cite{MorencyTutorial}. 

The uni-modal performance of individual modalities on any task may be significantly different~\cite{abavisani2019improving}; modalities whose individual performance is comparatively better (worse) are referred to as {\em stronger (weaker)} modalities~\cite{sayed2018cross}. For a given task, we can rank the available modalities according to their uni-modal performance. {\em Multi-modal fusion} methods combine the supplementary and complementary information provided by these modalities to improve performance compared to uni-modal methods~\cite{caridakis2007multimodal}\cite{tzirakis2017end}\cite{hu2019dense}. However, in general, most multi-modal fusion techniques require for the testing phase the simultaneous presence of all the modalities that were used during the model
training phase~\cite{baltruvsaitis2018multimodal}. This requirement becomes a severe limitation in case one or more sensors are missing or their signals are severely corrupted by noise during testing, unless such situations are explicitly handled by the modelling framework~\cite{mittal2020m3er}. Thus, it would be desirable to improve the testing performance of individual modalities using other modalities during training~\cite{abavisani2019improving}\cite{han2019emobed}\cite{garcia2019learning}. In particular, since the individual modalities corresponding to the same physical phenomenon might not perform equally well on the downstream task, our aim is to improve the uni-modal testing performance of a weaker modality by exploiting a stronger modality during training. 

To this end, we propose Stronger Enhancing Weaker (SEW), a framework for improving the testing performance of a weaker modality by exploiting a stronger modality {\em during the training phase  only}.
SEW is a supervised neural network framework for knowledge transfer across modalities. During training, the stronger modality serves as an auxiliary or guiding  modality that helps to create weaker-modality representations that are more discriminative than the representations obtained using uni-modal training for a classification or regression task. We achieve this by combining weaker-to-stronger modality translation and feature alignment with the stronger modality representations. This solution is based on the intuition that inter-modal translation can create intermediate representations that capture joint information between both modalities. Explicitly aligning the intermediate and the stronger modality representations further encourages the framework to discover components of the weaker modality that are maximally correlated with the stronger modality. Note that, after using SEW for training, the stronger modality is no longer required at testing. We show the effectiveness of our framework on the AVEC 2016 audio-visual continuous emotion recognition tasks and show that SEW improves the uni-modal performance of weaker modalities. 

\section{Related work}

Most works on multi-modal training for uni-modal performance enhancement are designed for tasks where the different modalities are different types of images. For example, multi-modal training using RGB and depth images improves the uni-modal performance for hand gesture recognition~\cite{abavisani2019improving}. This is achieved by forcing the modality-specific parts of the network to learn a common correlation matrix for their intermediate feature maps. Depth images also improve the test-time performance of RGB images for action recognition using an adversarial loss for feature alignment~\cite{garcia2019learning}. However, the modalities considered in these methods are images of equal size and the uni-modal networks have the same architecture, thus preventing their direct application to distinct modalities like audio, video and text, whose feature types and dimensionality differ. A few works have been proposed to address this problem~\cite{han2019emobed}\cite{pham2019found}\cite{zadeh2020foundations}. For sentiment analysis, a sequence-to-sequence network with cyclic translation across modalities generates an intermediate representation that is robust to missing modalities during testing~\cite{pham2019found}. A multi-modal co-learning framework improves the uni-modal performance of the text modality via training using audio, video and text modalities~\cite{zadeh2020foundations}. However, these methods primarily benefit the uni-modal performance of the text modality, which is the stronger modality for the task. In contrast, we aim to explicitly improve the weaker modality using the stronger modality during training. A joint audio-visual training and cross-modal triplet loss can be used to develop a face/speech emotion recognition system using multi-modal training~\cite{han2019emobed}. However, in such a system the weaker modality may degrade the performance of the stronger modality~\cite{han2019emobed}.

\section{Proposed Method}

\begin{figure}[t]
\centering
\includegraphics[width=9cm,height=9cm]{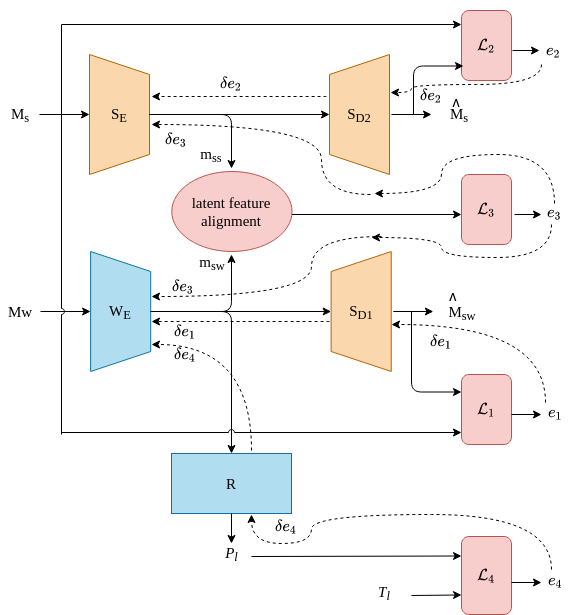}
\caption{The proposed SEW training framework. (M\textsubscript{S},M\textsubscript{W}) denotes a pair of stronger and weaker modality instances, S\textsubscript{E} and S\textsubscript{D2} represent intra-modal autoencoder, W\textsubscript{E} and S\textsubscript{D1} represent inter-modal translator, \^{M\textsubscript{S}} and \^{M\textsubscript{SW}} are the reconstructions of stronger modality from the encodings of stronger and weaker modalities respectively, R denotes the regressor/classifier connected to the inter-modal encoder, $T_{l}$ and $P_{l}$ stands for true and predicted labels, respectively, $m_{ss}$ and $m_{sw}$ represent the two latent representations, $\mathcal{L}_1$-$\mathcal{L}_4$ represent the 4 components of the total loss and $e_1$-$e_4$ are their respective error values. Dotted arrows represent the back-propagation of component error gradients. Only the blocks in cyan are retained during the deployment/inference phase.}
\label{fig:arch}
\end{figure}

In this section, we describe our proposed {\em Stronger Enhancing Weaker} (SEW), a supervised neural network framework, which uses jointly the stronger and the weaker modality representations during training to improve the testing performance of the weaker modality (Figure \ref{fig:arch}). The key concepts of our framework are inter-modality translation and feature alignment. These concepts are implemented using four main modules: an inter-modal translator, an intra-modal auto-encoder, a feature alignment module and a task-specific regressor or classifier. These modules are described next.

The inter-modal translator contains an encoder, W\textsubscript{E} and a decoder, S\textsubscript{D1}. The translator takes the features of the weaker modality, M\textsubscript{W}, as input and produces the features of the stronger modality, \^{M\textsubscript{SW}}, as output. The encoder of the inter-modal translator, creates intermediate representations, $m_{sw}$, that capture joint information across modalities. This is achieved by using a translation loss, $\mathcal{L}_{tr}$, between the true, M\textsubscript{S}, and the predicted, \^{M\textsubscript{SW}}, features of the stronger modality: 

\begin{equation}\label{eq:1}
\begin{aligned}
    \mathcal{L}_1 = \mathcal{L}_{tr}(M\textsubscript{S}, \hat{M}\textsubscript{SW}).
\end{aligned}
\end{equation}

W\textsubscript{E} is encouraged to discover components of the weaker modality that are inclined towards the stronger modality by increasing the alignment between $m\textsubscript{sw}$ and the representations of the stronger modality. For this purpose, we project the stronger modality features into the same latent space as $m\textsubscript{sw}$. We use an intra-modal auto-encoder to create stronger modality representations, $m\textsubscript{ss}$, of the same dimensionality as that of the inter-modal translator representations, $m\textsubscript{sw}$. To this end, we employ an auto-encoding loss, $\mathcal{L}_{ae}$, between true, M\textsubscript{S}, and predicted, \^{M\textsubscript{S}}, features: 
\begin{equation}\label{eq:2}
\begin{aligned}
    \mathcal{L}_2 = \mathcal{L}_{ae}(M\textsubscript{S}, \hat{M}\textsubscript{S}).
\end{aligned}
\end{equation}
For modality reconstructions, we use Mean-Square-Error (MSE) as $\mathcal{L}_{tr}$ and $\mathcal{L}_{ae}$ \cite{pham2019found}.

A feature alignment loss, $\mathcal{L}_{al}$, ensures that the intermediate representations of the inter-modal translator are maximally aligned to the stronger modality representations:
\begin{equation}\label{eq:3}
\begin{aligned}
    \mathcal{L}_3 = \mathcal{L}_{al}(m\textsubscript{ss},m\textsubscript{sw}).
\end{aligned}
\end{equation}
Following~\cite{sun2019multi}\cite{dumpala2019audio}, we use Canonical Correlation Analysis (CCA) for feature alignment, such that $\mathcal{L}_{al}$ = -CCA. CCA for deep neural networks, also known as Deep CCA or DCCA, is a method to learn complex nonlinear transformations of data from two different modalities, such that the resulting representations are highly linearly correlated \cite{andrew2013deep}. For a training set of size $p$, \begin{math} M_{s}  \in \R^{d_{1} \times p}  \end{math}
and \begin{math} M_{w}  \in \R^{d_{2} \times p}  \end{math}
are the input matrices corresponding to the stronger and the weaker modalities, respectively. $m_{ss} \in \R^{d \times p} $ and $m_{sw} \in \R^{d \times p} $ are the representations obtained by nonlinear transformations introduced by the layers in the encoders $ S_{E} $ and $ W_{E} $, respectively. Note that $ S_{E} $ and $ W_{E} $ bring the individual modalities with dimensions $d_1$ and $d_2$ into a common latent dimension $d$. If $ \theta_{es} $ and $ \theta_{ew} $ denote the vectors of all parameters of $ S_{E} $ and $ W_{E} $, respectively, then the goal of DCCA is to jointly learn parameters for both the views such that correlation, $(\rho)$, between $m_{ss}$ and $m_{sw}$ is as high as possible, i.e.,

\begin{equation}
\begin{split}
(\theta_{es}^{\star},\theta_{ew}^{\star}) & =  \operatorname*{arg\,max}_{\theta_{es},\theta_{ew}} \rho(m_{ss}, m_{sw}) \\
 & = \operatorname*{arg\,max}_{\theta_{es},\theta_{ew}} \rho(S_{E}(M_{S};\theta_{es}), W_{E}(M_{W};\theta_{ew})).
\end{split}
\end{equation}

If $ \bar{m}_{ss} $ and $  \bar{m}_{sw}  $ are the mean-centred versions of $m_{ss}$ and $m_{sw}$, respectively, then the total correlation of the top-K components of \begin{math} m_{ss} \end{math} and \begin{math} m_{sw} \end{math} is the sum of the top-K singular values of the matrix, \begin{math} T = \Sigma_{s}^{-1/2} \Sigma_{sw} \Sigma_{w}^{-1/2} \end{math}, in which the self (\begin{math} \Sigma_{s}, \Sigma_{w} \end{math}) and cross covariance (\begin{math} \Sigma_{sw} \end{math}) matrices are given by
\begin{equation} \label{eq:7}
    \Sigma_{sw} = \frac{1}{p-1} \bar{m}_{ss} \bar{m}_{sw}^T.
\end{equation}
\begin{equation} \label{eq:8}
    \Sigma_{s} = \frac{1}{p-1} \bar{m}_{ss} \bar{m}_{ss}^T + r_{1}I.
\end{equation}
\begin{equation} \label{eq:9}
    \Sigma_{w} = \frac{1}{p-1} \bar{m}_{sw} \bar{m}_{sw}^T + r_{2}I.
\end{equation}
where $r_1$ $>$ 0 and $r_2$ $>$ 0 are regularisation constants. We use the gradient of correlation obtained on the training data to determine \begin{math} (\theta_{es}^{\star},\theta_{ew}^{\star}) \end{math}.

Finally, the task-specific regressor or classification module, which takes the inter-modal translator representations as input, ensures the discriminative ability of the resulting latent space. We use a prediction loss, $\mathcal{L}_{pr}$, that operates on the true, $T_{l}$, and predicted task labels, $P_{l}$, as: 
\begin{equation}\label{eq:4}
\begin{aligned}
    \mathcal{L}_4 = \mathcal{L}_{pr}(T_{l}, P_{l}).
\end{aligned}
\end{equation}

The total training loss, $\mathcal{L}$ combines the four components:
\begin{equation}\label{eq:5}
\begin{aligned}
    \mathcal{L} = \alpha \mathcal{L}_1 + \beta \mathcal{L}_2 + \gamma \mathcal{L}_3 + \mathcal{L}_4,
\end{aligned}
\end{equation}
where $\alpha$, $\beta$ and $\gamma$ are  hyper-parameters. 
After training, all the components except the encoder, W\textsubscript{E}, and the regressor, R, are removed and the stronger modality is not required at the testing (deployment) phase.

\begin{table}[t!]
    \centering
    \caption{Unimodal results on the RECOLA development set. KEY - CCC: Concordance Correlation Coefficient, Acc: Binary Classification Accuracy (\%), geo: geometric, app: appearance. The best uni-modal results in arousal and valence are highlighted in bold.}
    \begin{tabular}{l|l|l|l|l|l|l}
    \specialrule{1.2pt}{0.2pt}{1pt}
            & \multicolumn{2}{c|}{Audio} & \multicolumn{2}{c|}{Video-geo} & \multicolumn{2}{c}{Video-app} \\
            \cmidrule(lr){2-3} \cmidrule(lr){4-5} \cmidrule(lr){6-7}
            & CCC         & Acc       & CCC           & Acc          & CCC           & Acc          \\
    \specialrule{1.2pt}{0.2pt}{1pt}
    Arousal &  \textbf{0.761}           &   \textbf{81.3}        &   0.482            &       66.4        &      0.492         &      69.1         \\
    \midrule
    Valence &     0.543        &       74.2      &   \textbf{0.643}            &    \textbf{81.9}           &     0.489          &      68.4         \\
    \specialrule{1.2pt}{0.2pt}{1pt}
    \end{tabular}
    \label{tab:unimodal}
\end{table}
\begin{table*}[t!]
    \centering
    \caption{Ablation results for SEW in terms of CCC and binary classification accuracy, Acc (\%). KEY - geo: geometric, app: appearance.}
    \begin{tabular}{l|c|c|c|c|c|c|c|c|c|c}
        \specialrule{1.2pt}{0.2pt}{1pt}
        & \multicolumn{4}{|c|}{Arousal} & \multicolumn{6}{c}{Valence}\\
         \midrule
         & \multicolumn{2}{c|}{video-geo(+audio)} & \multicolumn{2}{c|}{video-app(+audio)} & \multicolumn{2}{c|}{audio(+video-geo)} & \multicolumn{2}{c|}{video-app(+audio)} & \multicolumn{2}{c}{video-app(+video-geo)}\\ \hline
         & CCC & Acc & CCC & Acc & CCC & Acc & CCC & Acc & CCC & Acc \\ \hline
         SEW & \textbf{0.565} & \textbf{73.6} & \textbf{0.544} & \textbf{73.6} & 0.552 & 76.3 & \textbf{0.554} & \textbf{72.2} & \textbf{0.549} & \textbf{74.1} \\ 
         \hspace{4mm} -S$_{D2}$ & 0.532 & 71.1 & 0.519 & 71.5 & 0.486 & 72.4 & 0.539 & 68.7 & 0.540 & 74.0  \\ 
         \hspace{4mm} -CCA & 0.512 & 70.7 & 0.508 & 69.5 & 0.496 & 73.8 & 0.532 & 67.9 & 0.546 & 74.1 \\
         \hspace{4mm} -S$_{D1}$ & 0.514 & 71.0 & 0.523 & 71.0 & \textbf{0.556} & \textbf{76.3} & 0.514 & 67.8 & 0.505 & 69.3 \\
         \hspace{4mm} -(CCA \& S$_{D1}$) & 0.484 & 69.1 & 0.497 & 68.4 & 0.545 & 75.9 & 0.497 & 67.3 & 0.491 & 68.9 \\ 
         uni-modal & 0.482 & 66.4 & 0.492 & 66.1 & 0.543 & 74.2 & 0.489 & 68.4 & 0.489 & 68.4 \\
         \specialrule{1.2pt}{0.2pt}{1pt}
    \end{tabular}
    \label{tab:weaktostrong}
\end{table*}

\begin{table}[t!]
    \centering
    \caption{Performance comparison of SEW with other methods in terms of CCC. KEY - geo: geometric, app: appearance. Best and second best results are shown in bold and italics, respectively.}
    \begin{tabular}{l|l|l|l|l}
        \specialrule{1.2pt}{0.2pt}{1pt}
         & \multicolumn{2}{c|}{Arousal} & \multicolumn{2}{c}{Valence}\\
         \hline \hline 
         & video-geo & video-app & audio & video-app \\ \hline
         SVR + offset \cite{valstar2016avec} & 0.379 & 0.483 & 0.455 & 0.474 \\
         MTL (RE) \cite{han2017reconstruction} & 0.502 & 0.512 & 0.519 & 0.529 \\
         MTL (PU) \cite{han2017hard} & 0.508 & 0.502 & 0.506 & 0.468 \\
         DDAT (RE) \cite{zhang2018dynamic} & \textit{0.544} & 0.539 & 0.508 & 0.528 \\
         DDAT (PU) \cite{zhang2018dynamic} & 0.513 & 0.518 & 0.498 & 0.514 \\
         EmoBed \cite{han2019emobed} & 0.527 & \textbf{0.549} & \textit{0.521} & \textbf{0.564} \\
         SEW & \textbf{0.565} & \textit{0.544} & \textbf{0.552} & \textit{0.554} \\
         \specialrule{1.2pt}{0.2pt}{1pt}
    \end{tabular}
    \label{tab:compare}
\end{table}
\section{Validation}
In this section, we compare the performance of SEW with other uni-modal methods~\cite{valstar2016avec}\cite{han2017reconstruction}\cite{han2017hard}\cite{zhang2018dynamic} and a state-of-the-art cross-modal knowledge transfer method~\cite{han2019emobed}. We describe the dataset, the evaluation metrics, the details about the architecture and training, and  present an ablation study, which quantifies the contributions of different parts of SEW.

\subsection{Dataset and Evaluation Measures}
We use RECOLA, the AVEC 2016 emotion recognition dataset~\cite{valstar2016avec}, which contains audiovisual recordings of  spontaneous and natural interactions from 27 French-speaking participants. Continuous dimensional emotion annotations (in the range [-1,1]) in terms  of both {\em arousal} (level of activation or intensity) and {\em valence} (level of positiveness or negativeness) are provided with a constant frame rate of 40 ms for the first five minutes of each recording, by averaging the annotations from all annotators and also taking the inter-evaluator agreement into consideration~\cite{valstar2016avec}. The dataset is equally divided into three sets, by balancing gender, age, and mother-tongue of the participants with each set consisting of nine unique recordings, resulting in 67.5k segments in total for each part (training, development and test). Since, the test labels are not publicly available, we report the results on the development set. We have used the same audio and video features as the AVEC 2015 and 2016 baselines~\cite{valstar2016avec} for a fair comparison with the previous literature. These are 88-D extended Geneva Minimalistic Acoustic Parameter Set (eGeMAPS) features extracted using openSMILE, LGBP-TOP based 168-D video-appearance features and 49 facial landmarks based 632-D video-geometric features. It is to be noted that the dataset provides separate features for arousal and valence. As in~\cite{han2019emobed}\cite{valstar2016avec}, to compensate for the delay in annotation, we shift the ground-truth labels back in time by 2.4 s. This dataset is ideal for our objective, since the uni-modal performance of audio and video features varies considerably for arousal and valence, as reported in~\cite{han2019emobed} and confirmed by our experiments (see Table \ref{tab:unimodal}). As in the AVEC 2016 challenge, we use the Concordance Correlation Coefficient (CCC) (eq. \ref{eq:6}) as the primary evaluation metric:

\begin{equation} \label{eq:6}
    CCC = \frac{2\sigma_{xy}^{2}}{\sigma_{x}^{2} + \sigma_{y}^{2} + (\mu_{x} - \mu_{y})^2}, 
\end{equation}
where $x$ and $y$ are the true and the predicted labels, respectively, and $\mu_{x}$, $\mu_{y}$, $\sigma_{x}$, $\sigma_{y}$ and $\sigma_{xy}$ refer to their means, variances and covariance, respectively. We also evaluate the binary classification results by separating the true and predicted annotations into negative [-1,0] and positive (0,1] classes.

\subsection{Experiments}
In order to identify the stronger and weaker modalities, we first assess the unimodal performances of audio, video-geometric and video-appearance features for arousal and valence using a regressor similar to~\cite{han2019emobed}. The regressor consists of 4 single time-step GRU-RNN layers, each made up of 120 neurons, followed by a linear layer and trained using the MSE loss. The unimodal results thus obtained are shown in Table~\ref{tab:unimodal}. For arousal, the performance of audio surpasses both video-geometric and video-appearance features. For valence, the video-geometric features outperform audio and video-appearance features. Thus, we have 5 cases for cross-modal knowledge transfer from stronger to weaker modalities, namely video-geo(+audio) and video-app(+audio) for arousal and audio(+video-geo), video-app(+audio) and video-app(+video-geo) for valence, where the modality in parenthesis indicates the stronger modality. 

\subsection{Architecture and Training}
Because the proposed method combines multi-modal data with different characteristics, it was necessary to adjust various architectural parameters according to the characteristics of the given modalities rather than solving the problem using a generic model. Hence, the encoders and decoders of both inter-modal translator and intra-modal auto-encoder of the 5 multi-modal combinations vary from each other. 
Specifically, the encoder and decoder for each modality differ in terms of the number of linear layers and the number of neurons in each layer. Since the provided video-appearance features were already refined using PCA, we did not reduce the dimensionality further and used a single linear layer of size 168 for both its encoder and decoder. Thus, for all combinations that contain video-appearance features, the size of the latent dimension was 168. For all the rest, it was 128. The encoder and decoder for video-geometric features use linear layers of size [512, 256, 128] and [256, 512, 632], respectively with $\tanh$ activation between layers. For audio features, these were [108, 128] and [108, 88]. Note that 632 and 88 were chosen to match the dimensionality of the video-geometric and audio features, respectively. All the models were developed, trained and tested using PyTorch. We used the SGD optimiser with learning rate 0.001, momentum 0.7 and weight decay regularisation. The batch size was 32. The number of CCA components, K, was 10 in all the experiments. The contribution of each loss component was equally important: $\alpha=\beta=\gamma=1$.

\subsection{Results} 
Table~\ref{tab:weaktostrong} reports the results using the full SEW framework as well as after ablating individual components. The bottom row provides the uni-modal results for the weaker modalities for ease of comparison with the SEW results. Comparing the last 2 rows, we can see that the SEW-(CCA\&$S_{D1}$) results are close to the uni-modal results of the weaker modality. This is as expected since SEW-(CCA\&$S_{D1}$) contains only the W\textsubscript{E} and regressor with no interaction with the stronger modality.  
In all the 5 cases, SEW was able to improve the results from the uni-modal and SEW-(CCA\&$S_{D1}$) models both in terms of CCC and binary accuracy. For arousal video-geo(+audio) and video-app(+audio), removing the CCA based alignment causes a drop of 0.053 and 0.036, respectively in CCC and 2.9\% and 4.1\%, respectively in binary accuracy. The corresponding numbers for valence audio(+video-geo) are 0.056 and 2.5\%, respectively. These observations support the significance of the CCA based distribution alignment in the SEW framework. For valence video-app(+audio) and video-app(+video-geo), removing the decoder of the inter-modal translator causes a drop of 
0.040 and 0.044, respectively in CCC and 4.4\% and 4.8\%, respectively in binary accuracy, which indicates the effectiveness of the weaker-to-stronger modality translation. 

In Table~\ref{tab:compare}, we compare the best uni-modal results of SEW with the 4 most relevant uni-modal models~\cite{valstar2016avec}\cite{han2017reconstruction}\cite{han2017hard}\cite{zhang2018dynamic} and a cross-modal training method~\cite{han2019emobed} in terms of CCC. \cite{valstar2016avec} provides the baseline results on the RECOLA dataset for the AVEC 2016 challenge. The uni-modal baseline used an SVM based classifier on the individual features. SEW significantly outperforms the baseline uni-modal results for all the weaker modalities considered. Our method is also able to improve the uni-modal results for all the cases from~\cite{zhang2018dynamic}, which uses difficulty awareness based training and~\cite{han2017reconstruction}\cite{han2017hard} which uses multi-task learning. SEW outperforms EmoBed~\cite{han2019emobed} for arousal video-geo(+audio) and valence audio(+video-geo) by a margin of 0.038 and 0.031, respectively, in CCC. For arousal video-app(+audio), the performance of SEW and EmoBed are very close (0.544 and 0.549, respectively). However, for the valence video-app features, EmoBed outperforms SEW. The top and bottom rows of Table~\ref{tab:weaktostrong} show that SEW improves the uni-modal performance of the weaker modalities. Specifically, to the best of our knowledge, the best results to date on the uni-modal performance of arousal video-geometric features and valence audio features have been achieved by SEW.

\section{Conclusion}
We exploited the gap between the uni-modal performance of different modalities on a task to improve, using a stronger modality in a new training framework, the performance of the weaker modality. Our proposed framework, {\em Stronger Enhancing Weaker} (SEW), enables cross-modal knowledge transfer from the stronger to the weaker modality. The results of SEW on the RECOLA dataset for the task of continuous emotion recognition show its ability to improve the uni-modal performance of a weaker modality using a stronger modality during training. 

Future work includes applying the SEW framework to other tasks involving different features and modalities as well as extending SEW to cope with multi-modal sequential data. 

\vspace{.5cm}
\noindent{\bf{Acknowledgement}}. This research used QMUL's Apocrita HPC facility, supported by QMUL Research-IT. 
\bibliographystyle{IEEEbib}
\bibliography{strings,refs}

\end{document}